\documentclass[conference,letterpaper]{IEEEtran}
\IEEEoverridecommandlockouts
\usepackage{cite}      

\usepackage[dvips]{graphicx}  

\usepackage{amsmath,amssymb}   
\usepackage{graphicx}
\usepackage{color}
\usepackage[obeyspaces]{url}
\usepackage{algorithmic}
\usepackage{algorithm}
\usepackage{multirow}
\usepackage{float}
\usepackage{subfigure}
\usepackage{booktabs}
\usepackage{adjustbox}
\usepackage{array}
\usepackage[left=0.71in,top=0.94in,right=0.71in,bottom=1.18in]{geometry}
\usepackage{gensymb}

\usepackage{subfig}
\usepackage{mars}

\begin{document}
\title{\huge Fast 2D Map Matching Based on Area Graphs}

\author{\authorblockN{ Jiawei Hou, Haofei Kuang and S\"oren Schwertfeger}
\authorblockA{\textit{School of Information Science and Technology}\\
\textit{ShanghaiTech University}\\
\textit{Shanghai, China}\\
\textit{\{houjw, kuanghf, soerensch\}@shanghaitech.edu.cn}\\}%
}%
	%
%


\marsPublishedIn{Accepted for:} 		

\marsVenue{International Conference on Robotics and Biomimetics (ROBIO) 2019}

\marsYear{2019}

\marsPlainAutors{Jiawei Hou, Haofei Kuang and S\"oren Schwertfeger}


\marsMakeCitation{Fast 2D Map Matching Based on Area Graphs}{IEEE Press}

\marsDOI{\url{}}

\marsIEEE{}


\makeMARStitle

%
%
\maketitle
\begin{abstract}
	
We present a novel area matching algorithm for merging two different 2D grid maps. There are many approaches to address this problem, nevertheless, most previous work is built on some assumptions, such as rigid transformation, or similar scale and modalities of two maps. 
In this work we propose a 2D map matching algorithm based on area segmentation.
We transfer general 2D occupancy grid maps to an area graph representation, then compute the correct results by voting in that space. In the experiments, we compare with a state-of-the-art method applied to the matching of sensor maps with ground truth layout maps. The experiment shows that our algorithm has a better performance on large-scale maps and a faster computation speed.
\\
\end{abstract}


\section{Introduction}
\label{sec:introduction}

Map matching is an important and challenging task in the robotics field. Merging robot maps with maps in different modalities is beneficial for many robotics applications. 
For example, matching a robot sensor map with a known layout map could apply the semantic information from the layout map to the robot map, which is very helpful for human robot interaction \cite{kakuma2017alignment, georgiou2017constructing}. This can be extended to semantic localization, where semantic labels in one map can help with the localization of the robot in the current map. Furthermore, matching maps of different modalities, e.g., from robot maps to ground truth maps,
is a critical part of robot map evaluation \cite{schwertfeger2015Map}. Another important application of map matching is to help with the cooperation in multi-robot systems \cite{carpin2008fast}. 

Solving the autonomous map matching problem is interesting and challenging. The process of matching two different maps is similar to image registration. In practice, robot sensor maps can be treated as a picture. However, when an occupancy grid map is interpreted as an image, it has fewer and less descriptive features compared to a general digital image. Thus common image registration algorithms (e.g., feature-based methods) are not suitable to the map matching problem in robotics. 


There are three challenges of map matching problem: 1) Matching two different maps in different modalities with image registration only works with quite similar looking maps with big overlap \cite{birk2006merging, carpin2004stochastic, carpin2005map, carpin2008fast}. 2) These methods cannot deal with the scale problem, which can be present in maps from different sources. Some of these methods are based on the assumption of a rigid transformation. But large-scale maps sometimes are distorted due to problems with the SLAM algorithm, so non-rigid transformations would have to be applied.  
To address these challenges, topological maps have been used in map matching problem \cite{schwertfeger2013evaluation,schwertfeger2015Map,huang2005topological,wallgrun2010voronoi}. Shahbandi et\ al. \cite{shahbandi20192d} propose a state-of-the-art method which is based on region decomposition. This method can solve the different modality and scaling, and it also improves the performance with the distortion problem. 
3) The processing time in 2D map matching has to be considered. Especially for large scale maps the computation requirements can be intensive and thus prevent certain applications with real-time requirements, like multi-robot localization and map matching. 
 
In this work, we propose a graph matching algorithm which is based on our Area Graph  \cite{hou2019area}. Our method addresses the above mentioned problems. The presented algorithm can deal with maps from different modalities and it does not rely on rigid transformation assumption. Also, our method exhibits a fast computation speed.   Our algorithm has been implemented in C++ and tested on various maps in different modalities and scenarios. We also compare with the state-of-the-art method and the results show the high performance and efficiency of our algorithm, especially in the large-scale scenario.

This paper is organized as follows. In Section~\ref{sec:related_work}, we introduce the related work about map matching problems. Section~\ref{sec:overview} gives the overview of our algorithm. The details of area segmentation and matching will be discussed in Section~\ref{sec::method}. Section~\ref{sec:exp} shows the multiple results aiming to illustrate the performance and efficiency of our approach for different maps. We conclude our work in Section~\ref{sec:conclusion}.

\section{Related Work}
\label{sec:related_work}
In this section, we review related work concerning map matching. The map matching problem is related to the data association problem in robot mapping. Konolige et al. stated that map merging was a meaningful and challenging problem in the robotics field, but it did not receive as much attention as other robotics problems such as SLAM \cite{konolige2003map}. In the early stage of this research field, optimization is one of the most popular technique. Map matching is similar to image registration. Feature-based and optimization methods are mainstream methods, such as SIFT \cite{lowe2004distinctive} and Lucas-Kanade algorithm \cite{baker2004lucas,lucas1981iterative}. Maps can be considered as a picture, and the map matching could be considered as a special case of image registration. However, it offers less distinctive features than general digital pictures due to the self-similarity of the indoor environment.
These methods are easily affected to local minimal, particularly in absence of an initial guess. Carpin and Birk model the map matching problem as an optimization problem and then solve it through a stochastic search algorithm \cite{carpin2004stochastic, carpin2005map}. Their subsequent work improves the performance of their previous work through using mechanisms to detect failures and a more complex method to guide the search \cite{birk2006merging}.

Other methods could transform the input cartesian signal to a parametric space. This has the advantage of extracting the structure of the maps in the parametric space. Carpin proposed a method using the Hough transform to model the search space and decompose the transformation into translation and rotation estimation for merging occupancy maps in multi-robot systems \cite{carpin2008fast}. Because of the decomposition, these methods are often fast, non-iterative, and deterministic. However, these approaches are limited to some assumptions: the input maps being merged have been built using the same scale and just work best on maps which are in same modality. These methods are also limited to rigid transformations, and would not satisfy the matching in a large-scale environment because of affection of distortions.

A common approach to address the problem is to model the input maps to an abstract representation and then search on the similarity of instances. Huang et al. proposed an algorithm which is based on graph matching and image registration to merge partial maps with embedded topological maps \cite{huang2005topological}. Through a similar approach, Wallgrün et al. propose a map matching algorithm using a graph matching based on the Voronoi graph \cite{wallgrun2010voronoi}. Schwertfeger et al. developed a method for map matching which is used for map quality assessment in an automated process \cite{schwertfeger2015Map, schwertfeger2013evaluation} based on the matching of vertices, which takes their descriptor and the similarity of their neighbors into account. Saeedi et al. use the topological structure to find the relative transforms between two maps through Radon transform and find translation through an edge matching methods \cite{saeedi2012efficient,saeedi2014group,saeedi2014map}. 

For fusing prior maps and robot maps, the state-of-the-art approach to the map matching problems must allow the input maps with different modalities. Georgiou et al. state that computing the correspondence between abstract human readable maps with a map which is constructed from a robot's sensors is desired to improve the effectiveness of human-robot cooperation \cite{georgiou2017constructing}. A related topic is the matching of paths in a topological map, which can then be used for map matching \cite{schwertfeger2016matching}. For finding the correct matching between different modalities of the maps, Shahbandi et al. propose a decomposition-based algorithm to match 2D spatial maps \cite{shahbandi20192d}. Through region decomposition, they abstract maps into 2D arrangements which explicitly represent both the boundaries and the region of the open-spaces. It makes the method capable of interpreting maps in different modalities and scale. 
Their method achieves state-of-the-art performance in different modality map matching tasks, but in order to avoid missing the correct results, they exhaust the search space, which makes their algorithm slow. To address these problems, we propose an area graph based method to utilize the feature of the areas (e.g., corridor or doors).

\section{Overview}
\label{sec:overview}

\begin{figure}[tb]
	\centering
	\includegraphics[width=0.95\linewidth]{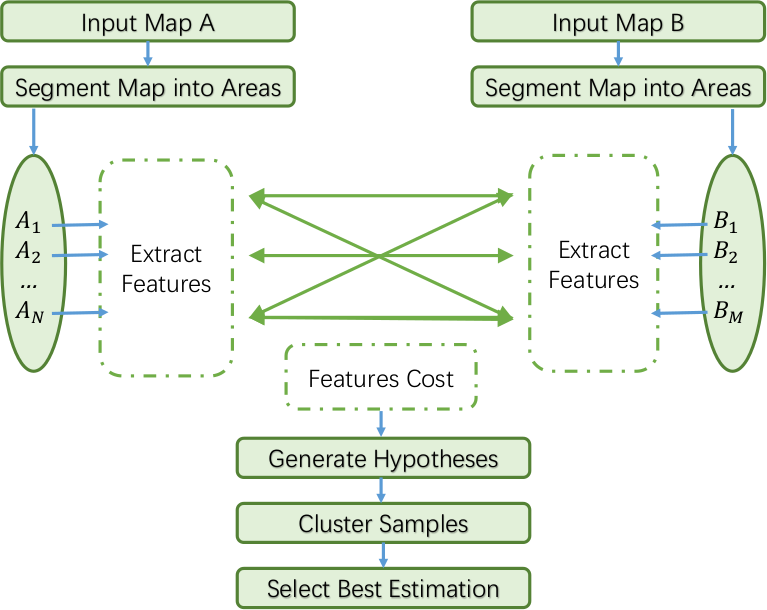}
	\caption{The pineline of our method.}
	\label{fig:pipeline}
\end{figure}

Given two overlapping maps of the same environment, the workflow of our algorithm is as follows:
\begin{enumerate}
	\item Segment the given 2D maps as areas, in this work we generate the Area Graph \cite{hou2019area} as the segmentation solution. 
	\item Compute the features for areas segmented from the map. The features used for matching include the size of the area, \textit{passage distance} and \textit{convex hull longest distance} of the areas, where passage distance is the distance between a pair of passages of the same area, which are basically connections between different areas. 
	\item Compute the area matching cost for each pair of areas by multiplying the feature cost with a weight vector $w$, and record the matching cost in an adjacency matrix $M_{cost}$.
	\item Find a list of pairwise areas that are considered as matched with each other. For the matched area pairs, estimate the rotation between them and save the angles as candidates.
	\item Cluster the rotations candidates to vote the range in which the correct rotation between two maps is located, which is called \textit{best cluster}.
	\item Traverse the samples in the best cluster and compute their overlap area to choose the best transformation.
\end{enumerate}

In Section~\ref{sec::method}, we introduce these modules in detail. The overview of our algorithm is shown in Fig. \ref{fig:pipeline}. 

\section{Methodology}
\label{sec::method}
\begin{figure}[tpb]
	\centering
	\includegraphics[width=0.49\linewidth]{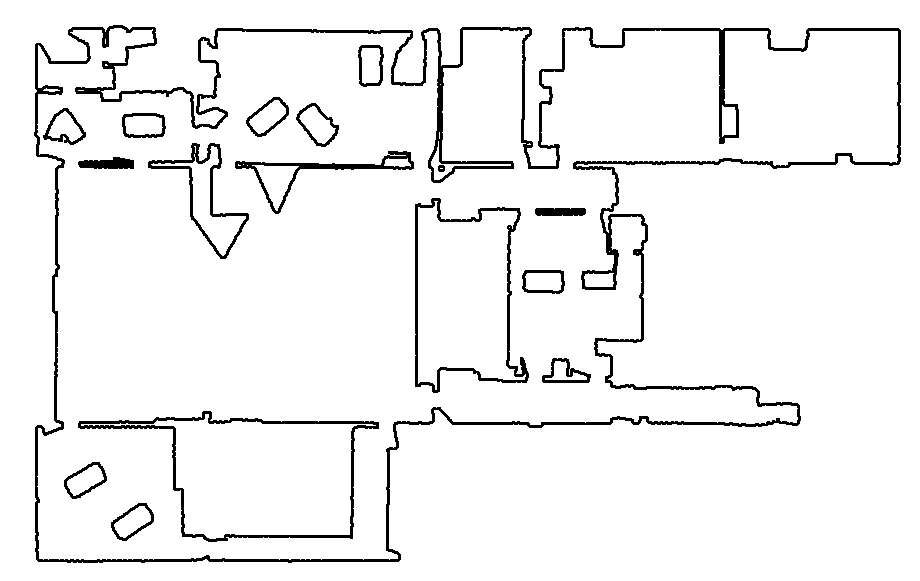}
	\includegraphics[width=0.49\linewidth]{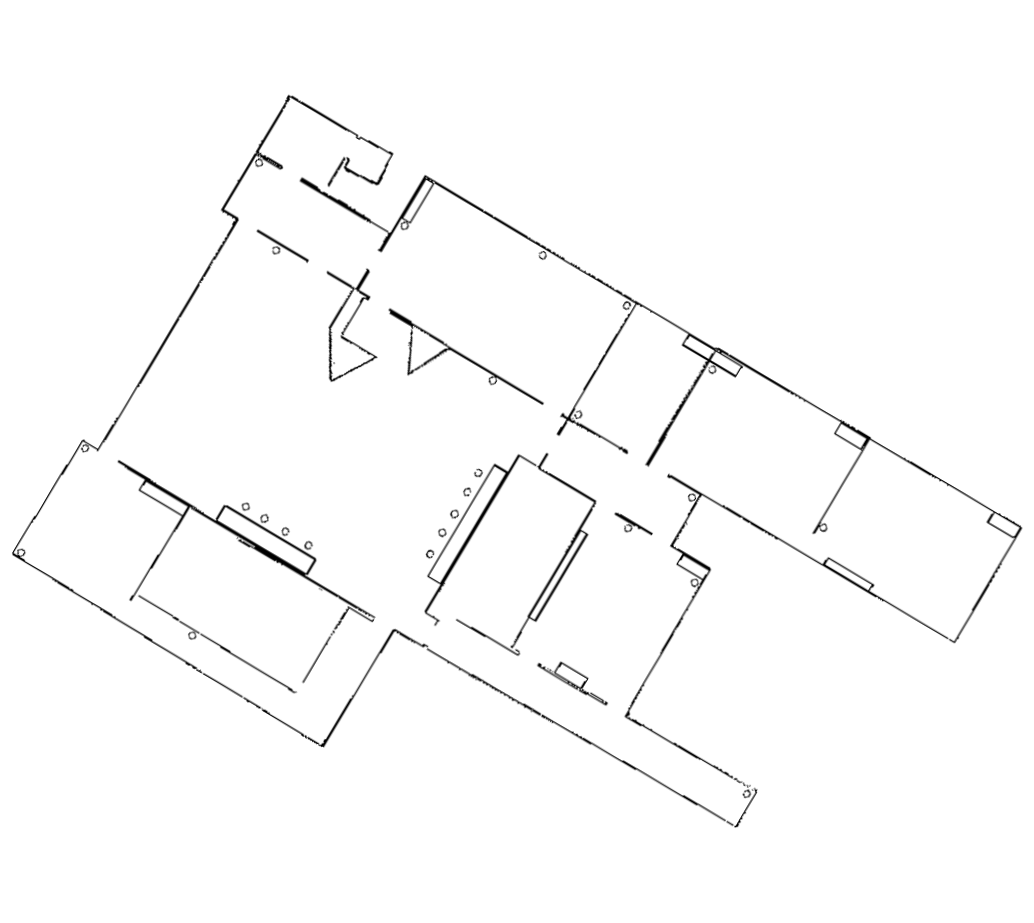}	
	\caption{The maps to be matched. The left map is binarized from a scanned map and the right map is a corresponding artificial map, both of which are from Bormann's dataset \cite{bormann2016room}.}
	\label{fig:maps}
\end{figure}
We use the two maps shown in Fig. \ref{fig:maps} as an example to explain our method. 


\subsection{Brief Introduction of Area Graph}
\label{sec::AreaGraph}
\begin{figure}[tpb]
	\centering
	\includegraphics[width=0.49\linewidth]{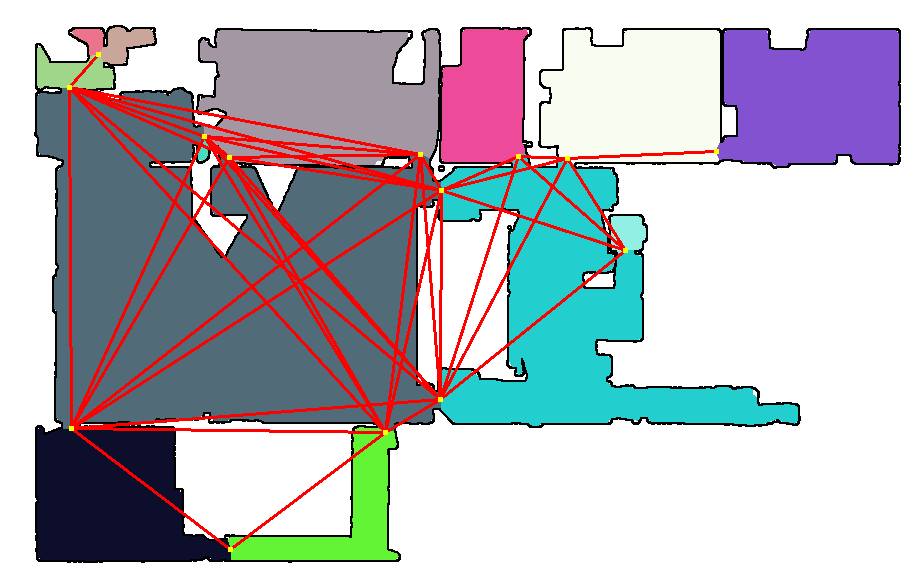}
	\includegraphics[width=0.49\linewidth]{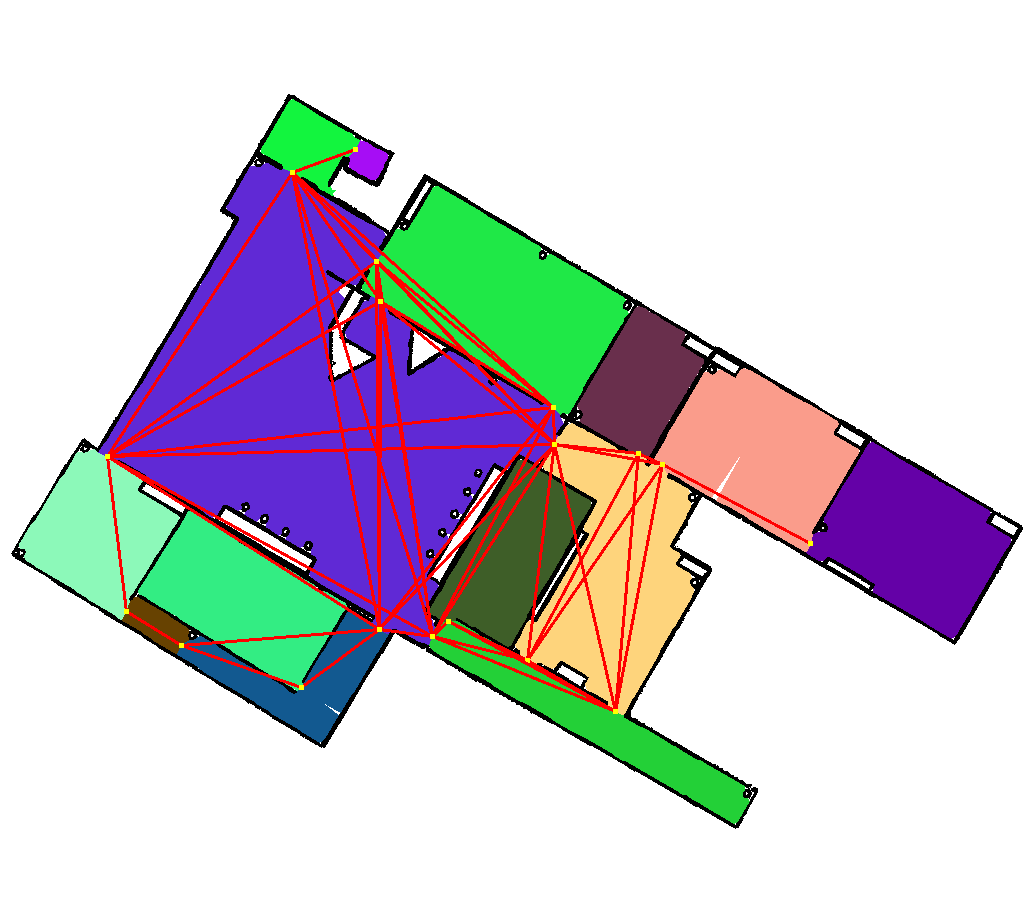}	
	\caption{The maps are divided into different areas, which are represented with different colors. The passages points are highlighted as yellow points, and we connect the passages belonging to the same area with red lines. The passage distance features are computed from the lengths of the red lines.}
	\label{fig:seg}
\end{figure}

In this paper, we segment the given map by generating the Area Graph, which is introduced in our previous work \cite{hou2019area}. The Area Graph is generated from a Topology Graph, which in turn is generated from the Voronoi Diagram of the 2D grid map. Additionally the $\alpha$-shape algorithm is used to detect and merge open areas such as rooms. More details can be found in \cite{hou2019area}. 
Due to the dependence of segmentation, our matching algorithm works with the assumption that the maps can be segmented to an enough number of areas which are not seriously over-segmented. As an example, the segmentation of the maps given in Fig. \ref{fig:maps} is visually shown in Fig. \ref{fig:seg}. 

\begin{figure}[tb!]
	\centering
		\label{fig:area_normal_zp}
		\includegraphics[width=0.49\linewidth ]{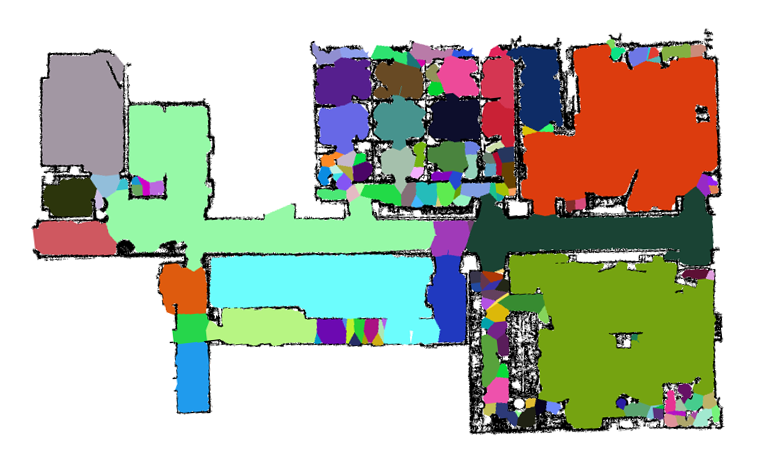}
		\label{fig:area_our_zp}
		\includegraphics[width=0.49\linewidth ]{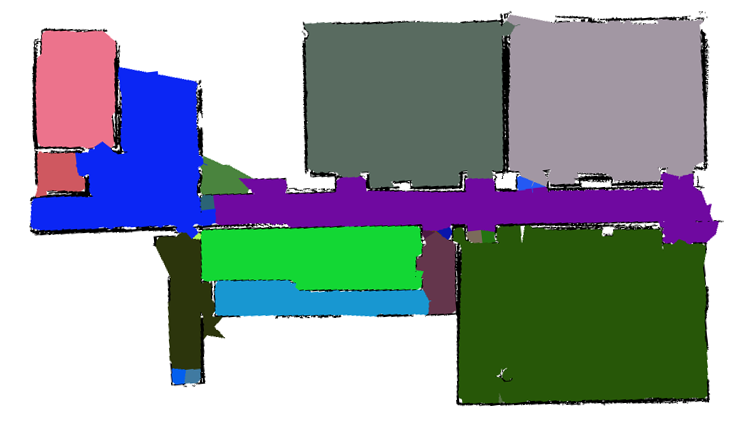}
	\caption{Furniture free 2D grid mapping: Left: Area Graph of a normal map with furniture. Right: A furniture free map from the same dataset with its Area Graph.}
	\label{fig:areaGraph_zp}
\end{figure}

To address the over-segmentation problem, we have developed an algorithm for the generation of furniture free maps \cite{he2019furniture}. It is based on a mapping robot with two 3D laser scanners: a horizontally scanning and a vertically scanning sensor. The robot additionally has 9 color cameras, an IMU and odometry. All sensors are fully synchronized \cite{Chen2019Towards} and calibrated \cite{Chen2019Heterogeneous}. Fig. \ref{fig:areaGraph_zp} shows the results of the Area Graph algorithm on maps from the same dataset, once with a normal 2D grid mapping algorithm and once with the furniture free mapping method. 

\subsection{Feature Extraction and Cost Computation}
\label{sec::feature_extraction}
For each area in the Area Graph, we compute its {\it size}, {\it passage distance} and {\it convex hull longest distance} as features to match areas between maps. 

\subsubsection{Area Size}
The regions in the map are represented as bounded polygons, and the area size is the polygons' most basic geometric property having meaningful recognition. 
Note the fact that the unit used for area size is $ m^2 $, while that for the other two features are $m$, and the total cost is a weighted sum of the three feature costs. In order to maintain the linearity of the weighted sum, it is reasonable to calculate the area size cost $c_a$ with the square roots of the area sizes:
\begin{align}
\label{equ:DA2}
c_a^{ij}=\frac{|\sqrt{a_i}-\sqrt{a_j}|}{\sqrt{\max(a_i,a_j)}}.
\end{align}
Here the area cost $ c_a $ between two areas is the normalized distance between the area size of the $ i $th region in the first map, donated as $a_i$, and the area of the $ j $th region in the second map, donated as $a_j$.

\subsubsection{Passage Distance}
\begin{figure}[tpb]
	\centering
	\includegraphics[width=0.49\linewidth]{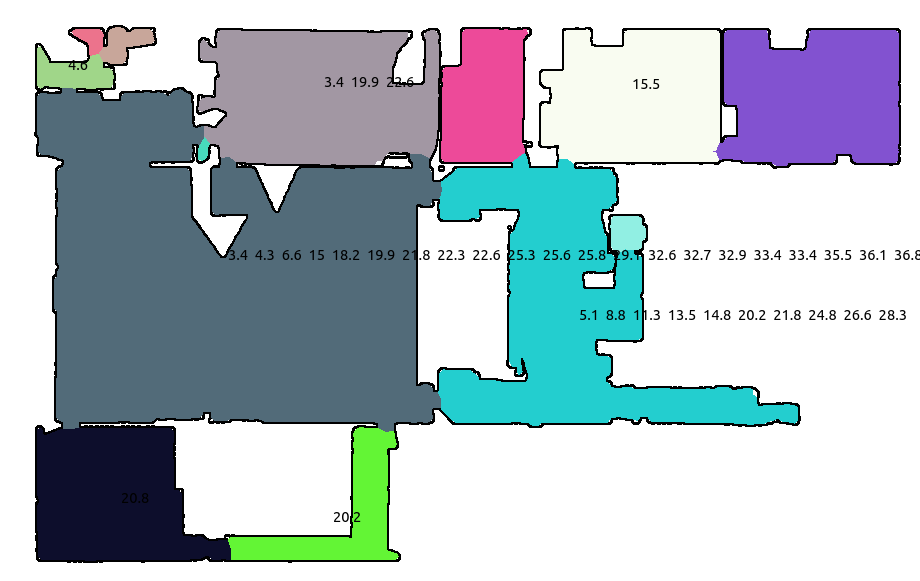}
	\includegraphics[width=0.49\linewidth]{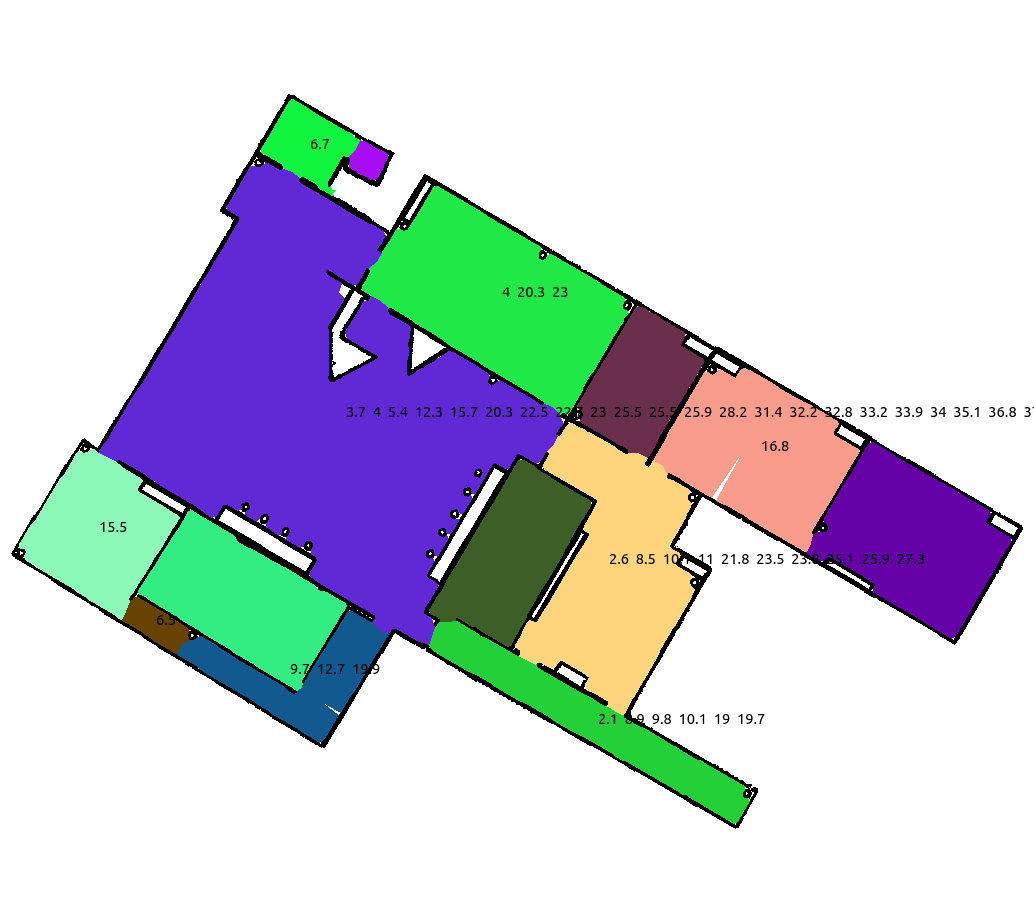}	
	\caption{The black numbers show the distance between each pair of passages in the regions. In the left example, the second left-top room with three doors has three passage distances close to the three passage distances involved in the corresponding region of the right map.}
	\label{fig:passageDistance}
\end{figure}

We notice that the distance between doors and the distance between junctions in corridors are fixed, which are called passages. Therefore, we developed the idea that compares the distance between passages as a feature to match areas. In Fig. \ref{fig:seg} all the passages in the same area are connected with red lines. We calculate the distance between all passages in each area. The distance between a pair of passages is donated as $ pd_n^i $, where $i$ is the area ID and this is the $n$th passage distance in the area $i$, e.g., the $N$ passage distances included in area $a_i$ are $\{ pd_1^i, \dots, pd_N^i \}$. An example is shown in Fig. \ref{fig:passageDistance}.

Then the minimal difference between passage distances in area $i$ and passage distances in the area $j$ is the passage distance cost $c_p^{ij}$ between the areas $i$ and $j$:
\begin{align}
\label{equ:DP}
c_p^{ij}=\min_{n,m} \frac{|pd_n^i - pd_m^j |}{\max(pd_n^i , pd_m^j)}.
\end{align}

\subsubsection{Convex Hull Longest Distance}

\begin{figure}[tpb]
	\centering
	\includegraphics[width=0.49\linewidth]{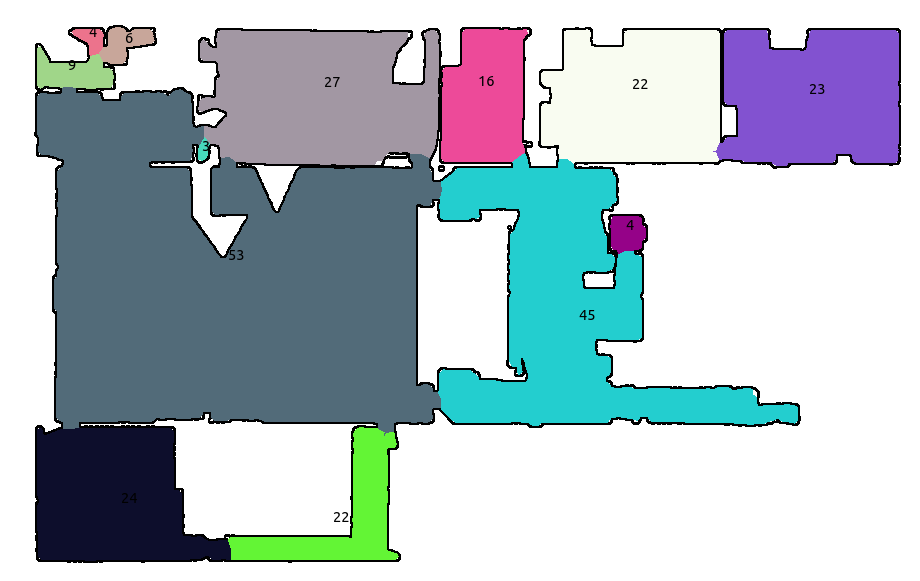}
	\includegraphics[width=0.49\linewidth]{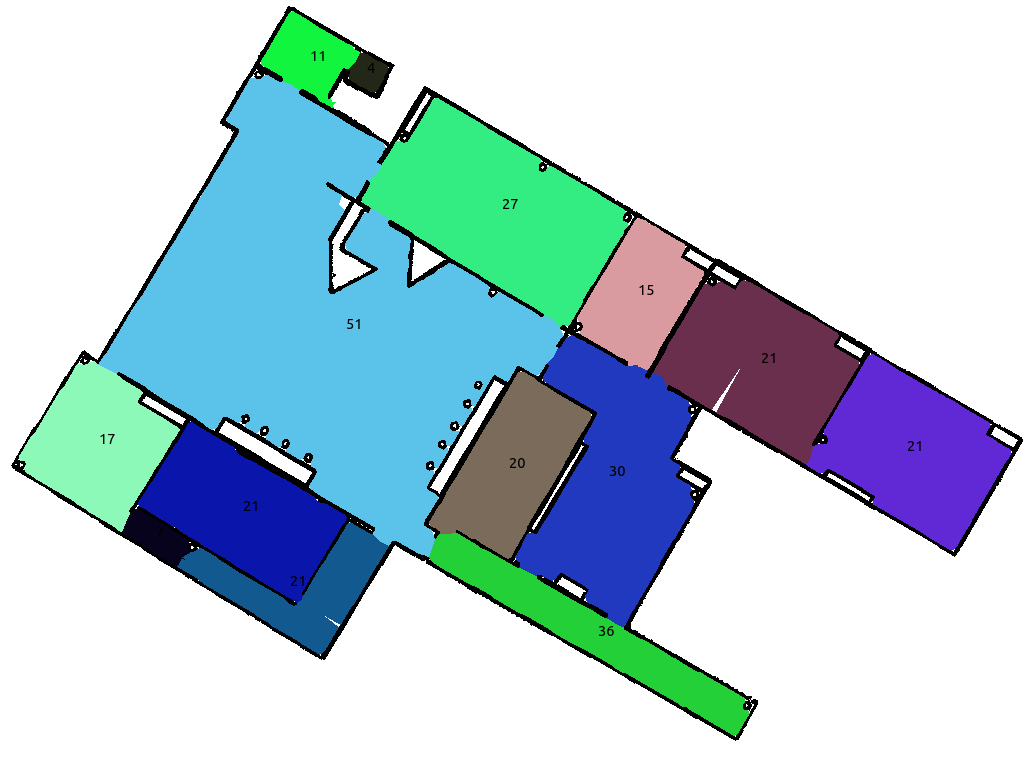}	
	\caption{The black numbers show the longest distance in each convex hull of regions. }
	\label{fig:longgestline}
\end{figure}

There are some areas that only contain one passage. As a result, the passage distance of these regions cannot be obtained. An example of the convex hull longest distance features of each area is shown in Fig. \ref{fig:longgestline}. We obverse that for corresponding areas the their convex hull  longest distance are very close, even better than the accuracy of the feature passage distance. However, since this feature can be affected by the scan integrity of the map, while the passage distance is robust, we only use this feature for the regions that cannot obtain passage distance.
The longest distance cost between two regions is computed as
\begin{align}
\label{equ:DL}
c_l^{ij}=\frac{|ld_i - ld_j |}{\max(ld_i, ld_j)}.
\end{align}

\subsection{Area Matching}
The goal of this step is to find a list of potential matching areas according to the features described above.
\subsubsection{Total Cost between Areas}
\label{subsec::total}
Each area in map $A$ is compared to all the $m$ areas in map $B$. For each pair to compare, we compute the feature distance for the three features described above, and save in a row vector.
Then the total matching cost between this area pair is computed by multiplying the feature cost vector and a weight vector as a weighted sum of all features cost. The matching costs between the area $i$ to all $m$ areas in another map are recorded in $C_i$.
\[
C_i =
\begin{bmatrix}
d_{11}       & d_{12} & d_{13}  \\
d_{21}       & d_{22} & d_{23} \\
\vdots	&\vdots & \vdots \\
d_{m1}       & d_{m2} & d_{m3} 
\end{bmatrix}
\begin{bmatrix}
w_{1} \\ w_{2}\\  w_{3}
\end{bmatrix}
=
\begin{bmatrix}
c_{i1} \\ c_{i2}\\ \vdots \\ c_{im}
\end{bmatrix}
\]
where $d_{jf} $   is the $f$th feature cost between the area $i$ and the area $j$ in another map for $j=1, \dots, m ~\text{and}~ f=1,2,3$. In Section~\ref{sec:exp1}, the setting of the weight parameters will be discussed.


\subsubsection{Potential Matching Pairs}
For each area in map $A$, we record its $k$ (e.g., k=3) most similar areas, i.e. the areas with the lowest matching cost to it, in map $B$, and vice versa to each area in map $B$.  Area $ i $ and $ j $ are considered as mutual matching areas if and only if area $ i $ is one of the k nearest matches of area $ j $ and area $ j $ is also one of the k nearest matches of area $ i $. We record all the mutual matching pairs in a matched pairs list, which will be used to generate transformation hypotheses.

\subsection{Transformation Estimation}
\label{sec:transform} 

\begin{figure}[tb]
	\centering
	\includegraphics[width=0.98\linewidth]{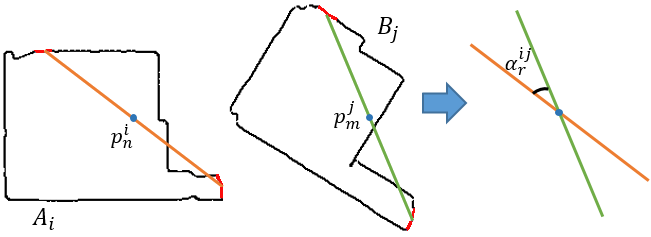}
	\includegraphics[width=0.98\linewidth]{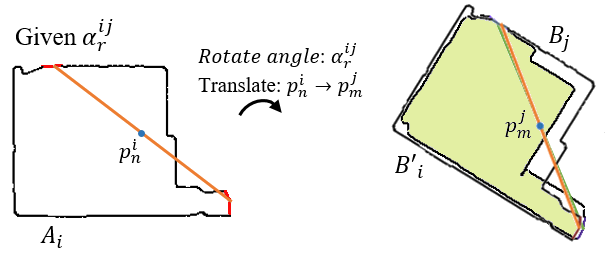}
	\caption{(a) Compute the transformation between two areas. (b) Transform the area $A_i$ with the given angle and the midpoints of its matched segment and the one of its matched area $B_j$. The passages are highlighted in red.}
	\label{fig:IVD1}
\end{figure}
\subsubsection{Rotation between matched areas}
Given a pair of matched areas ($A_i, B_j$), the transformation between $A_i$ and $B_j$ is obtained from their matched passage distance segments or longest lines segments in their convex hulls. 
The rotation is computed as the difference of the angles between the two segments, donated as $\alpha_{r}^{ij}$, with the midpoints of the segments $p_n^i$ and $p_m^j$ as the rotation center, to transform the area $A_i$ in the map A to the map B as $B_i'$. Then we calculate the overlap between the areas $B_i'$ and $B_j$ and use it as a criterion for judging whether the two areas are matched correctly. 
This process is illustrated in Fig. \ref{fig:IVD1}.
Then the rotation angles between correctly matched areas are recorded in a list as rotation hypotheses. 

\subsubsection{Cluster the Rotations}
We cluster these rotation hypotheses to vote the range in which the correct rotation between two maps is located.
The clustering algorithm is based on the angular distance between the angles. The clustering process starts with setting the first rotation in the list as the center of a new cluster. Given another rotation in the list, if its distance from all centers is more than the set angle threshold (e.g., 3 degrees), then we create a new cluster centered on it, otherwise, add it into the cluster whose center is closest to it and recalculate the center of the cluster. Finally, we iteratively adjust the clusters to ensure that, for each cluster, the distance from the rotations in the clusters to the center does not exceed the set threshold.
After the clustering is completed, the range of clusters that have the most angle hypotheses, donated as $Cluster_{max}$, is considered to be the range containing the correct map rotation.

\subsubsection{Search for the Best Rotation}
Given the best rotation cluster $Cluster_{max}$, we traverse its samples. For each sample $(\alpha_r^{ij}, (p_n^i, p_m^j))$, we transform all the areas in map A of matching area pair to match their corresponding areas in map B and sum up their overlap percentage.
For a pair of matched regions $(i, j)$, whose area values are $a_{i}$ and $a_{j}$, their overlap area percentage $OP$ is computed by 
$$ OP= \frac{Overlap\_Area}{\min(a_{i}, a_{j})}. $$
Then we select the sample obtains highest overlap sum as our final transformation estimation between the given maps.


\section{Experiments}
\label{sec:exp}
\begin{figure}[tpb]
	\centering
	\includegraphics[width=0.49\linewidth]{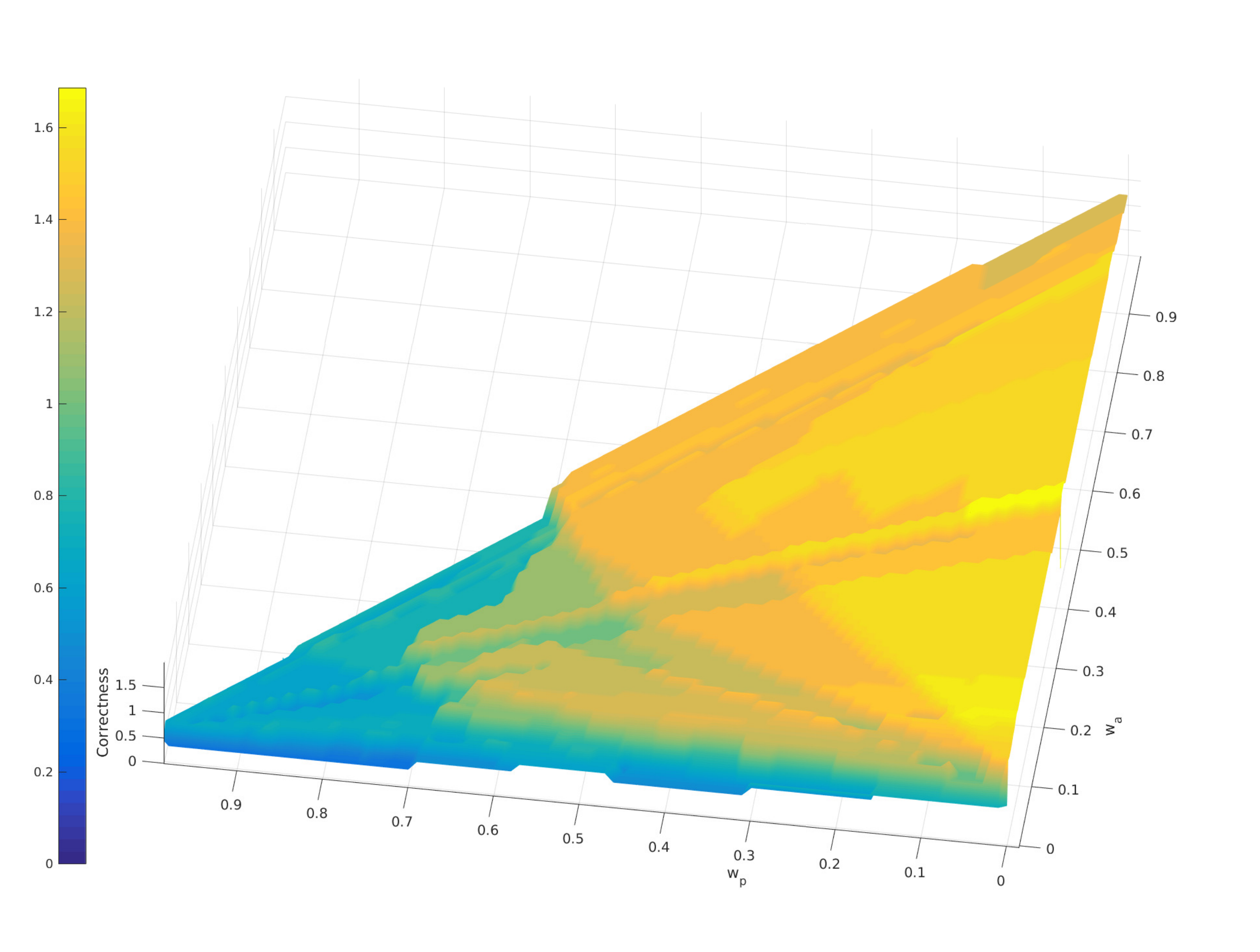}
	\includegraphics[width=0.49\linewidth]{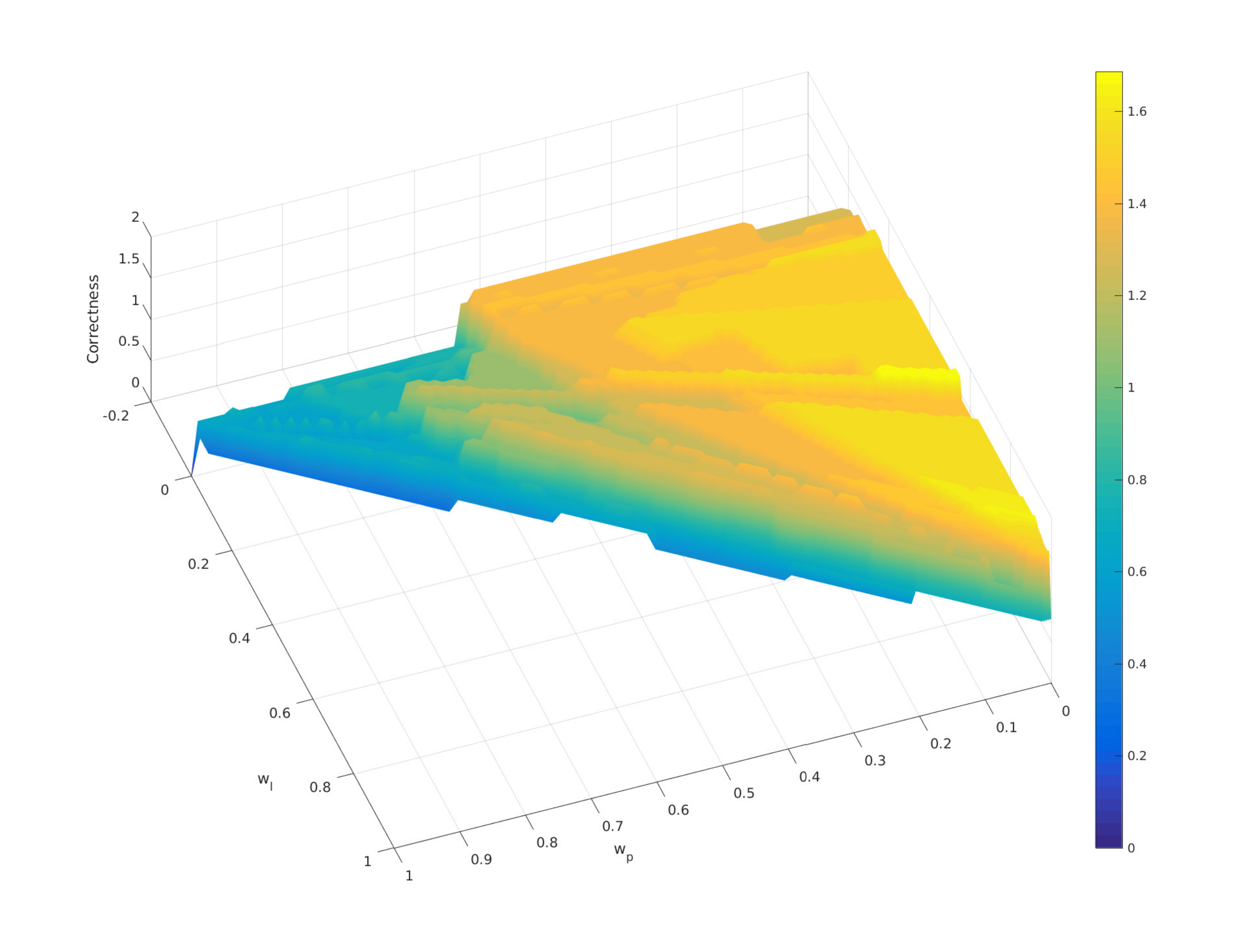}	
	\caption{A heatmap to show the correct matching number for different $w$ vectors. The yellow region represent the high correctness.}
	\label{fig:w_max}
\end{figure}
\subsection{Selection of $w$}
\label{sec:exp1}
As we mentioned in Section~\ref{subsec::total}, the matching cost between two areas is obtained by multiplying the feature vector with the weight vector $w=[w_a~ w_p~ w_l]^\top$. Both area value and passage distance are essential to deciding the matchings. We want to know the weighting ratio of these three features in making decisions. Therefore, this experiment is done to select a good $w$. We run this experiment on Bormann's dataset \cite{bormann2016room}, which includes large indoor environment maps and satisfies the assumption that the maps contain enough rooms.

In order to facilitate the search, and since only the weight ratio is important, we let the weight vector $w$ have an element sum of 1 and set the search step size to 0.01 for each element. 
To find a good ratio between $w_a$ and $w_p$, we use the $w_a$ as X-axis, the $w_p$ as Y-axis and the matching correctness of the best cluster as Z-axis to draw a heatmap. Also, we use the $w_p$ as X-axis and the $w_l$ as Y-axis to draw another heatmap. Both heatmaps are shown in Fig. \ref{fig:w_max}.
The correctness was obtained by dividing the correct area matchings in the best cluster by the hand-made ground truth.


According to Fig. \ref{fig:w_max}, we find some regulation for the set of $w$ to get a high correctness: $w_a$ and $w_p$ are positively correlated; $w_p$ and $w_l$ are negatively correlated; the lower the $w_p$ was, the higher the correctness was obtained. Therefore, we set $w=[0.1~ 0.1~ 0.8]^\top$ to run the comparison in the next experiment, which satisfies the regulation.


\subsection{Comparison with the State-of-the-Art Method}
We chose Shahbandi and Magnusson's work \cite{shahbandi20192d} to compare with our method. It is a state-of-the-art matching method, which also depends on segmentation. We observe that their method performed better on the maps with strong self-similarity for the reason that it exhausts the search space to find the best solution, which makes their algorithm unsuitable for matching large environments.  However, our algorithm is difficult to match the areas through feature matching for high self-similarity maps. More seriously, these maps can fool the overlap calculations, which lead to the wrong hypotheses selection. Their dataset included maps with high self-similarity and the maps with few rooms, contrary to our assumption, so we did not use their dataset. 

\begin{figure*}[b]
	\centering
	\renewcommand{\thesubfigure}{\space} \makeatletter
	\renewcommand{\@thesubfigure}{\space}
	\renewcommand{\p@subfigure}{\thefigure} \makeatother
	
	\rotatebox{90}{\hspace{.5cm}freiburg 101} 
	\subfigure{ \includegraphics[width=0.15\linewidth]{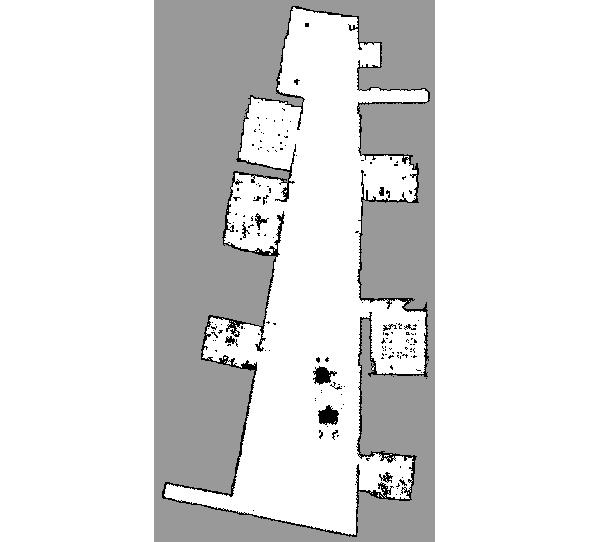}} \hspace{0.3cm}
	\subfigure{ \includegraphics[width=0.15\linewidth]{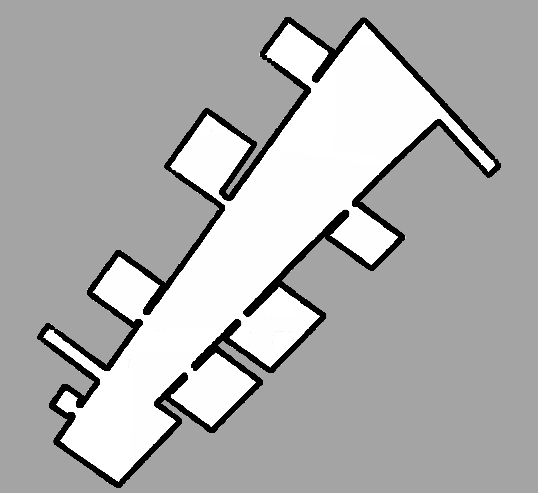}} \hspace{0.3cm}
	\subfigure{ \includegraphics[width=0.15\linewidth]{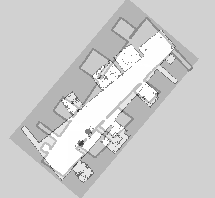}} \hspace{0.3cm}
	\subfigure{ \includegraphics[width=0.15\linewidth]{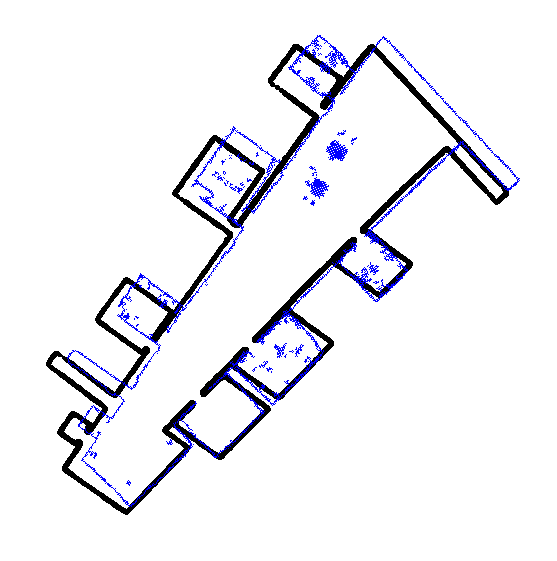}} \hspace{0.3cm}
	\subfigure{ \includegraphics[width=0.15\linewidth]{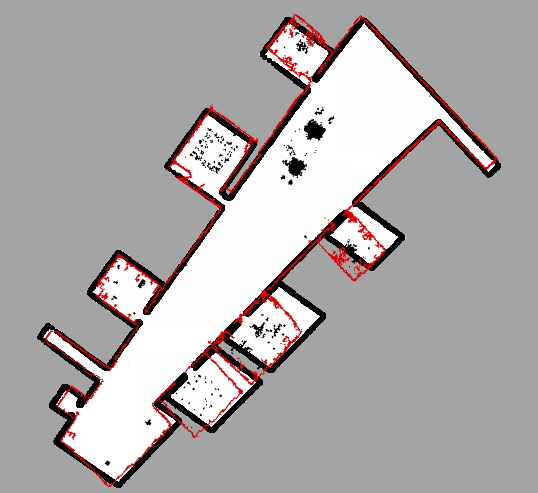}} \\
	\vspace{0.1cm}

	\rotatebox{90}{\hspace{1.cm}intel} 
	\subfigure{ \includegraphics[width=0.15\linewidth]{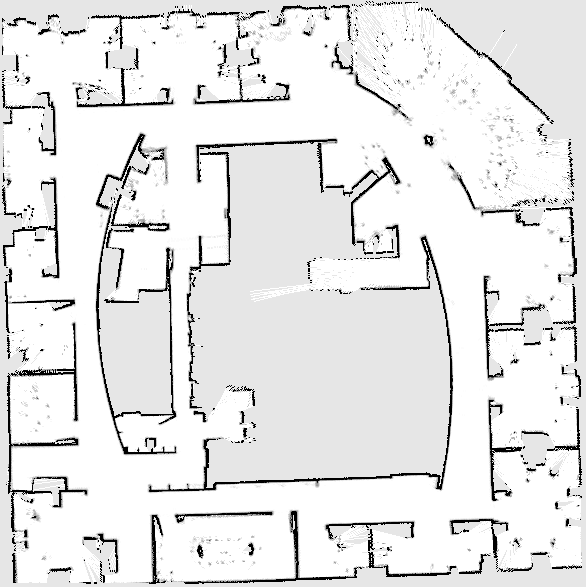}} \hspace{0.3cm}
	\subfigure{ \includegraphics[width=0.14\linewidth]{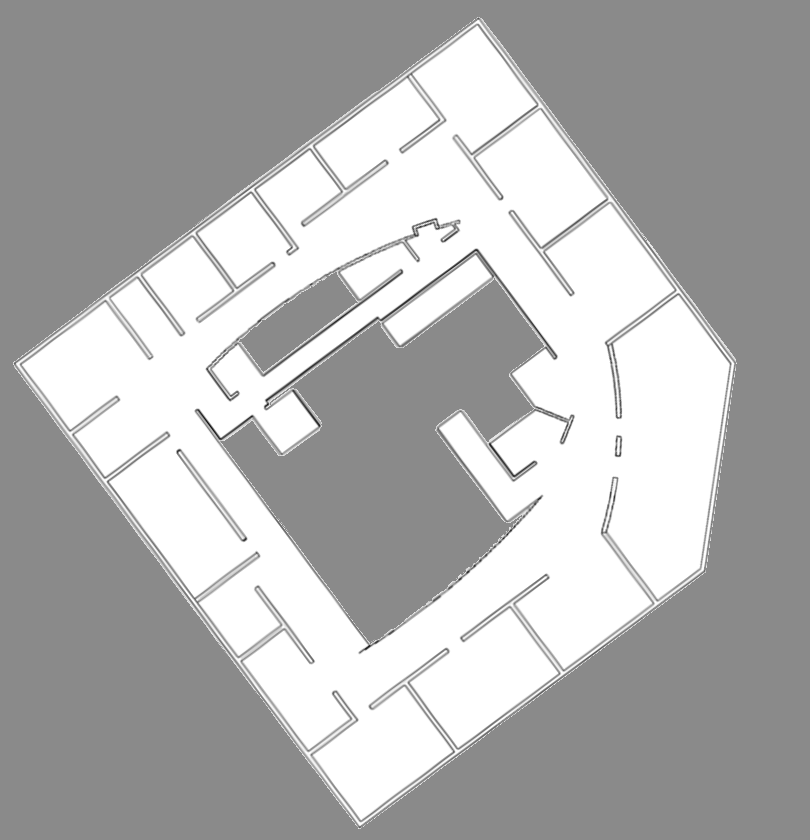}} \hspace{0.3cm}
	\subfigure{ \includegraphics[width=0.15\linewidth]{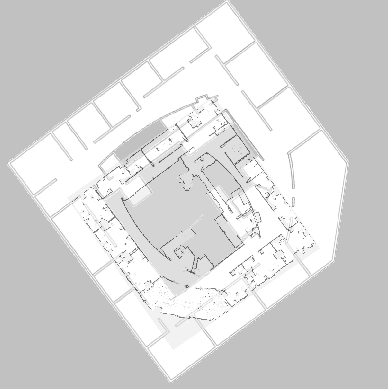}} \hspace{0.3cm}
	\subfigure{ \includegraphics[width=0.15\linewidth]{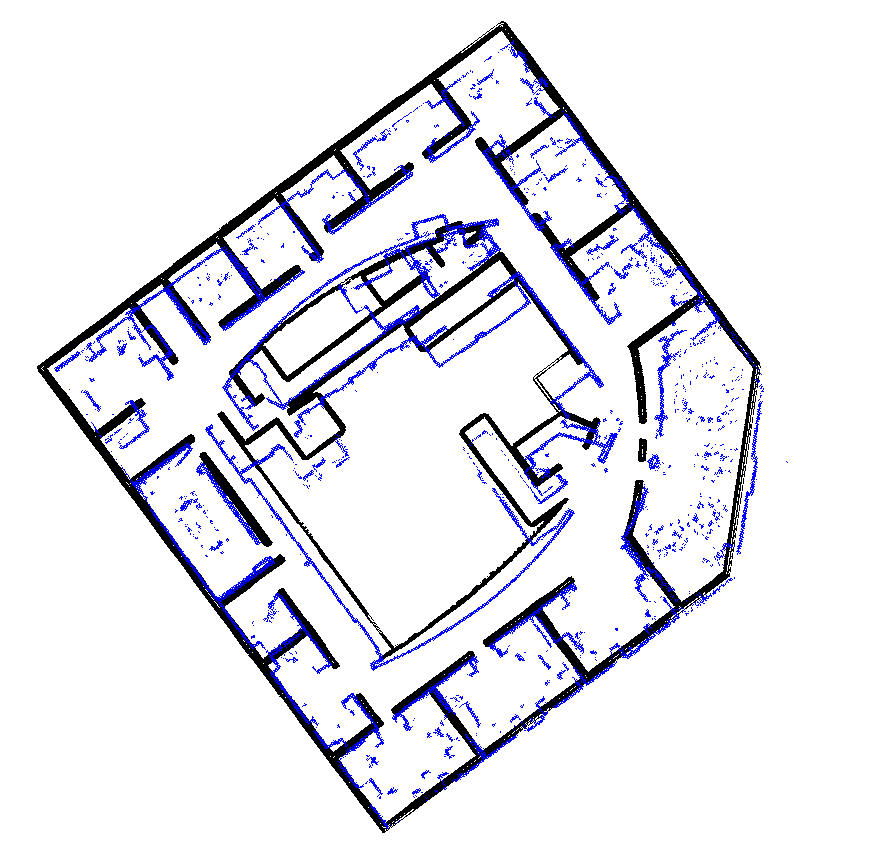}} \hspace{0.3cm}
	\subfigure{ \includegraphics[width=0.15\linewidth]{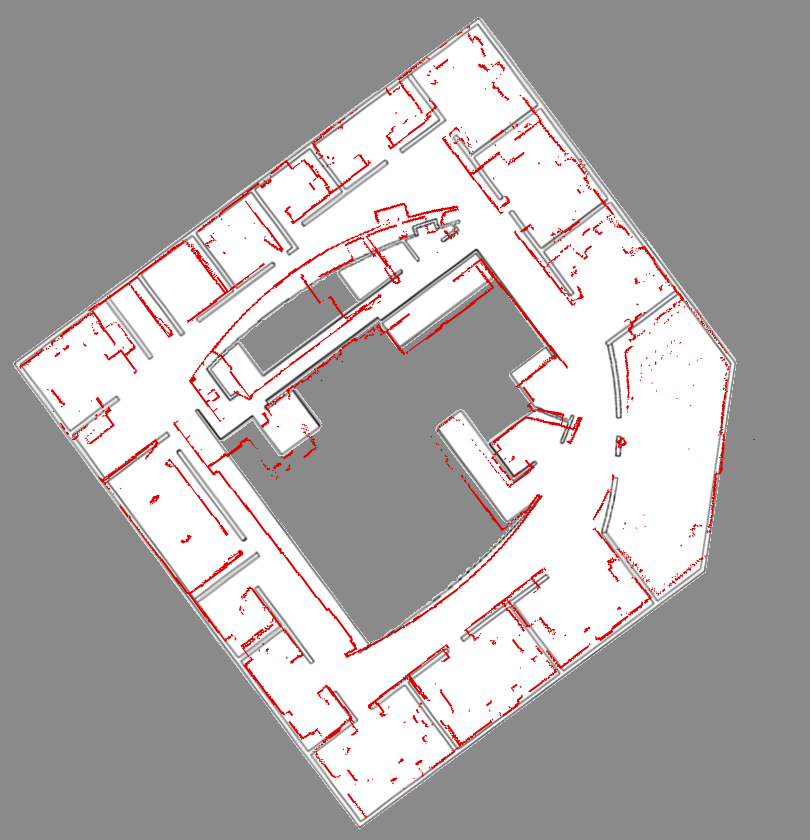}} \\
	\vspace{0.1cm}

	\rotatebox{90}{\hspace{1.cm}lab\_a} 
	\subfigure{ \includegraphics[width=0.15\linewidth]{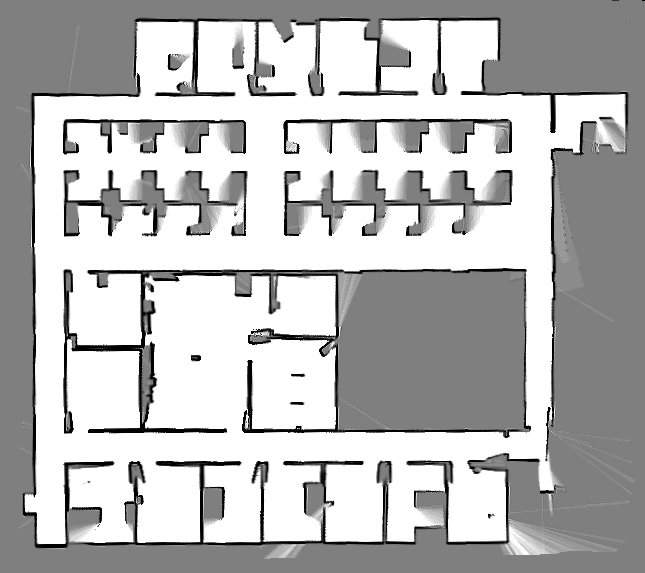}} \hspace{0.3cm}
	\subfigure{ \includegraphics[width=0.14\linewidth]{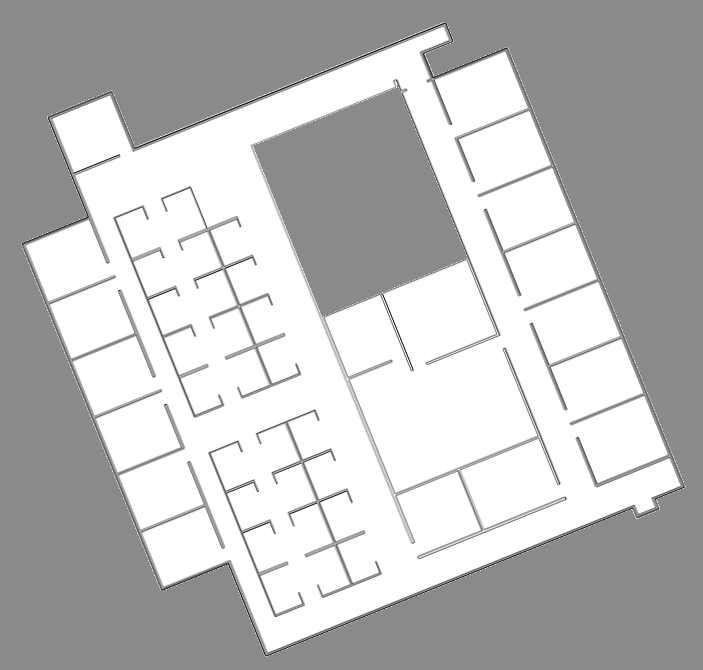}} \hspace{0.3cm}
	\subfigure{ \includegraphics[width=0.15\linewidth]{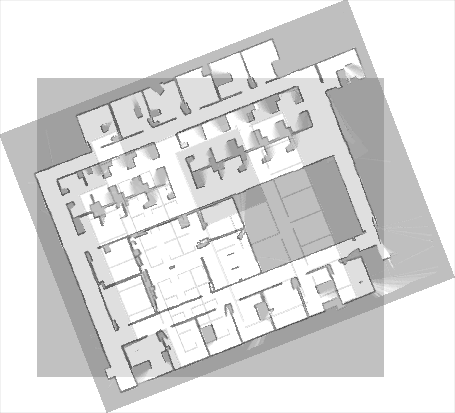}} \hspace{0.3cm}
	\subfigure{ \includegraphics[width=0.15\linewidth]{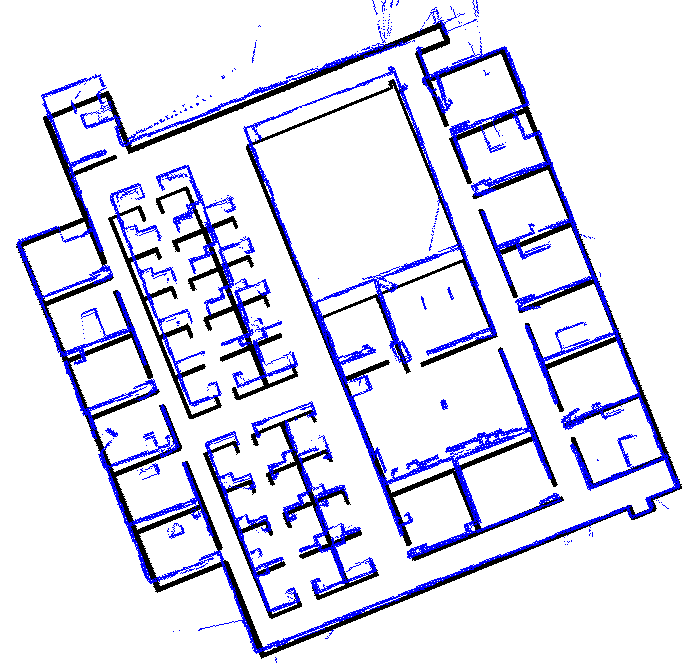}} \hspace{0.3cm}
	\subfigure{ \includegraphics[width=0.15\linewidth]{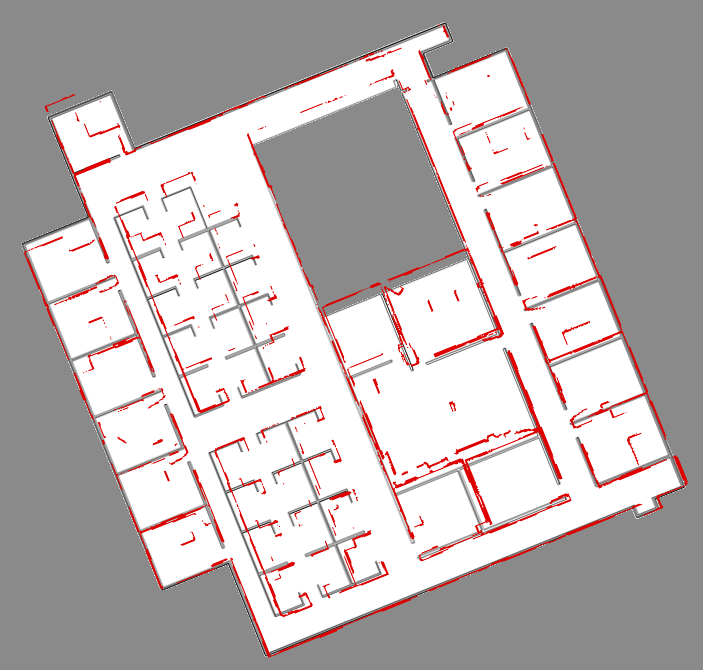}} \\
	\vspace{0.1cm}

	\rotatebox{90}{\hspace{.5cm}lab\_c} 
	\subfigure{ \includegraphics[width=0.15\linewidth]{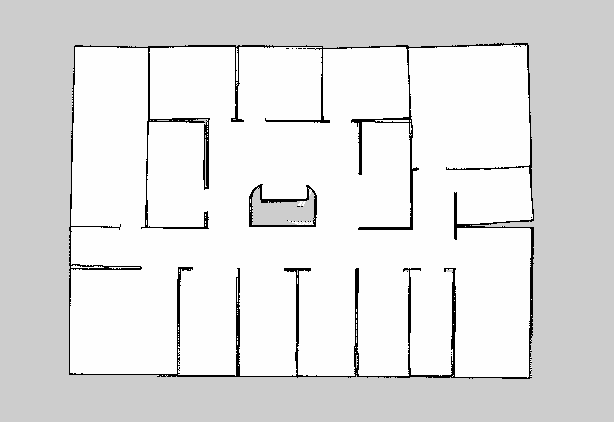}} \hspace{0.3cm}
	\subfigure{ \includegraphics[width=0.14\linewidth]{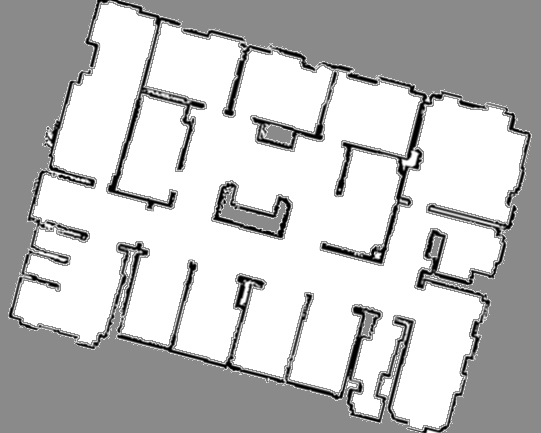}} \hspace{0.3cm}
	\subfigure{ \includegraphics[width=0.15\linewidth]{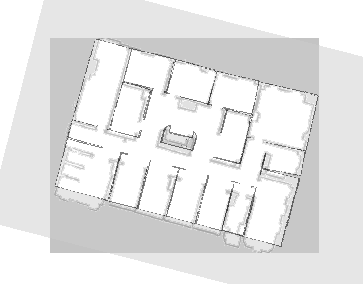}} \hspace{0.3cm}
	\subfigure{ \includegraphics[width=0.15\linewidth]{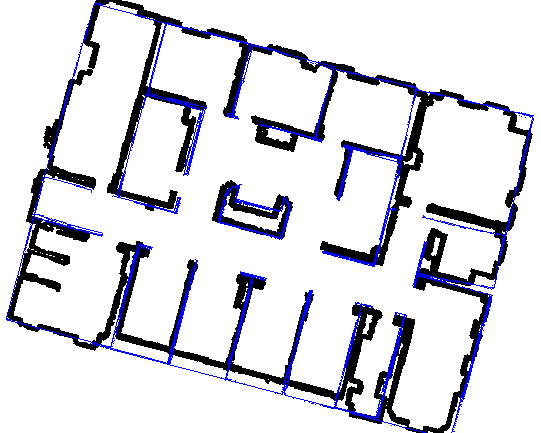}} \hspace{0.3cm}
	\subfigure{ \includegraphics[width=0.15\linewidth]{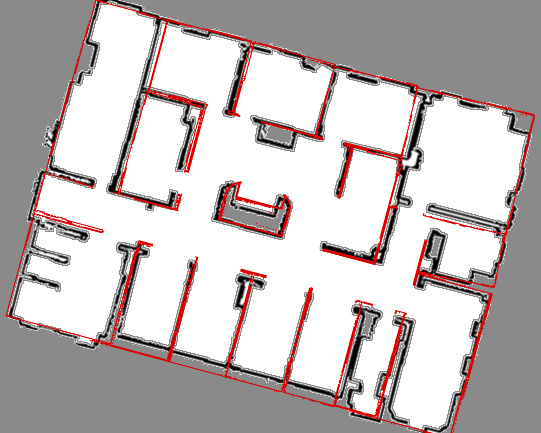}} \\
	\vspace{0.1cm}

	\rotatebox{90}{\hspace{.4cm}lab\_d} 
	\subfigure{ \includegraphics[width=0.15\linewidth]{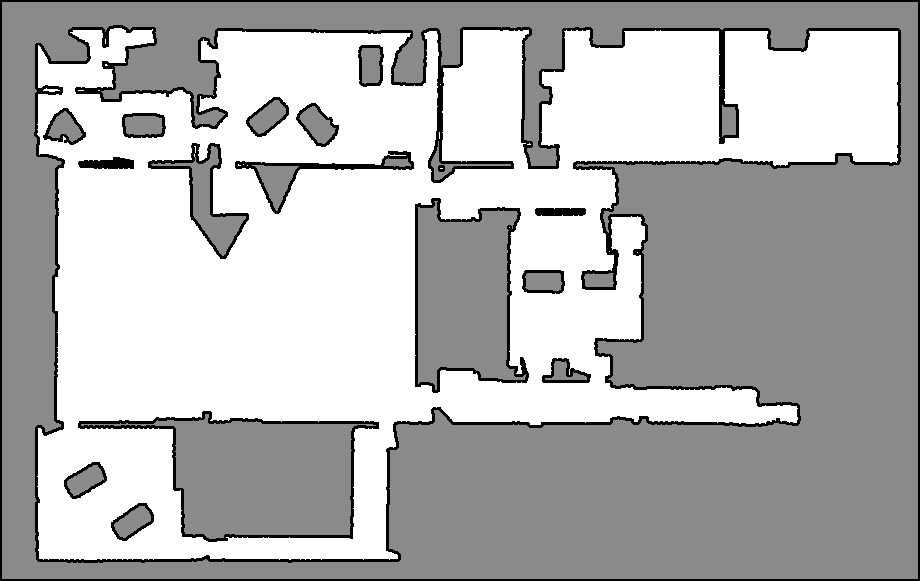}} \hspace{0.3cm}
	\subfigure{ \includegraphics[width=0.14\linewidth]{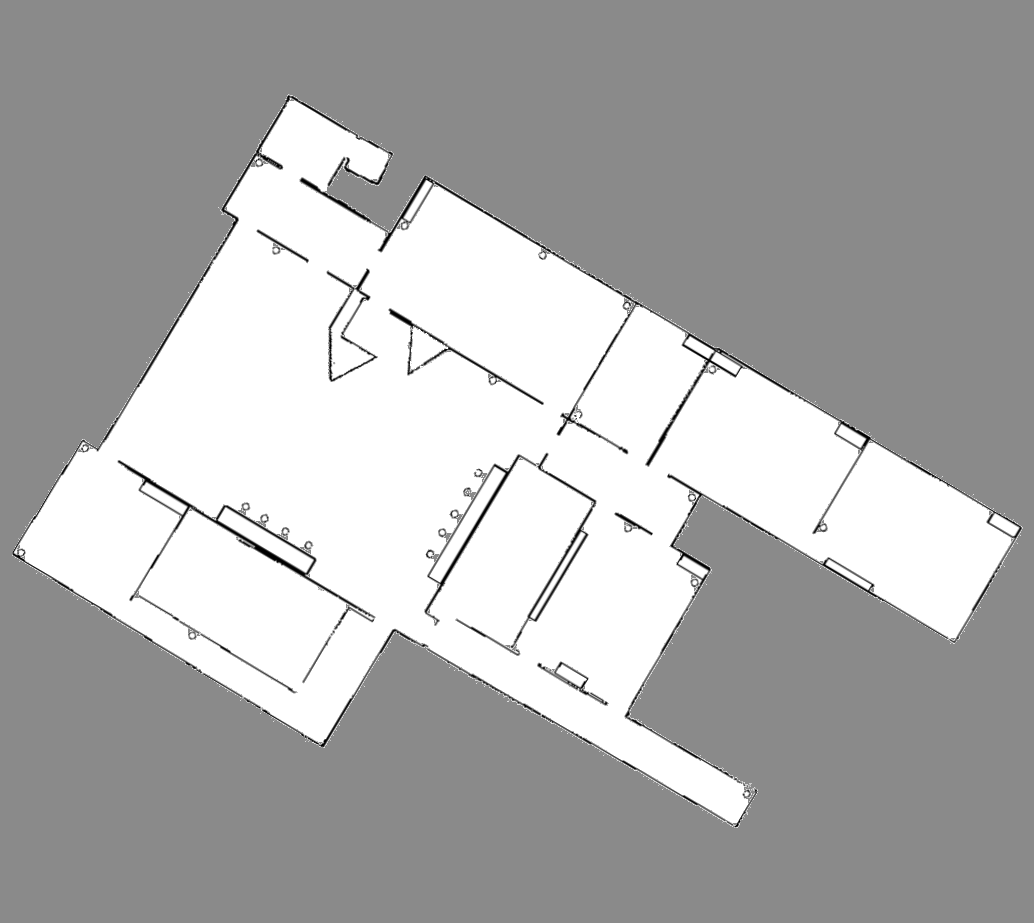}} \hspace{0.3cm}
	\subfigure{ \includegraphics[width=0.14\linewidth]{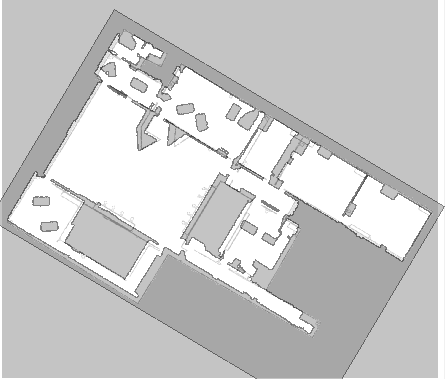}} \hspace{0.3cm}
	\subfigure{ \includegraphics[width=0.15\linewidth]{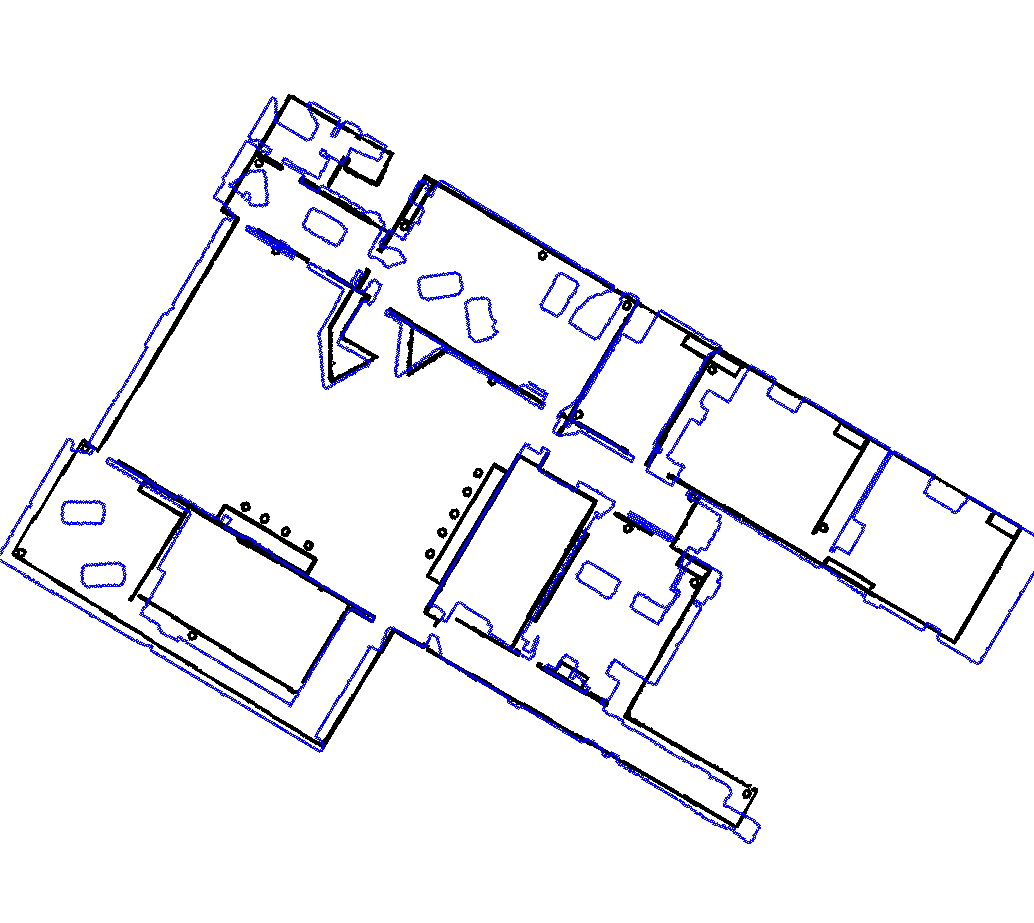}} \hspace{0.3cm}
	\subfigure{ \includegraphics[width=0.15\linewidth]{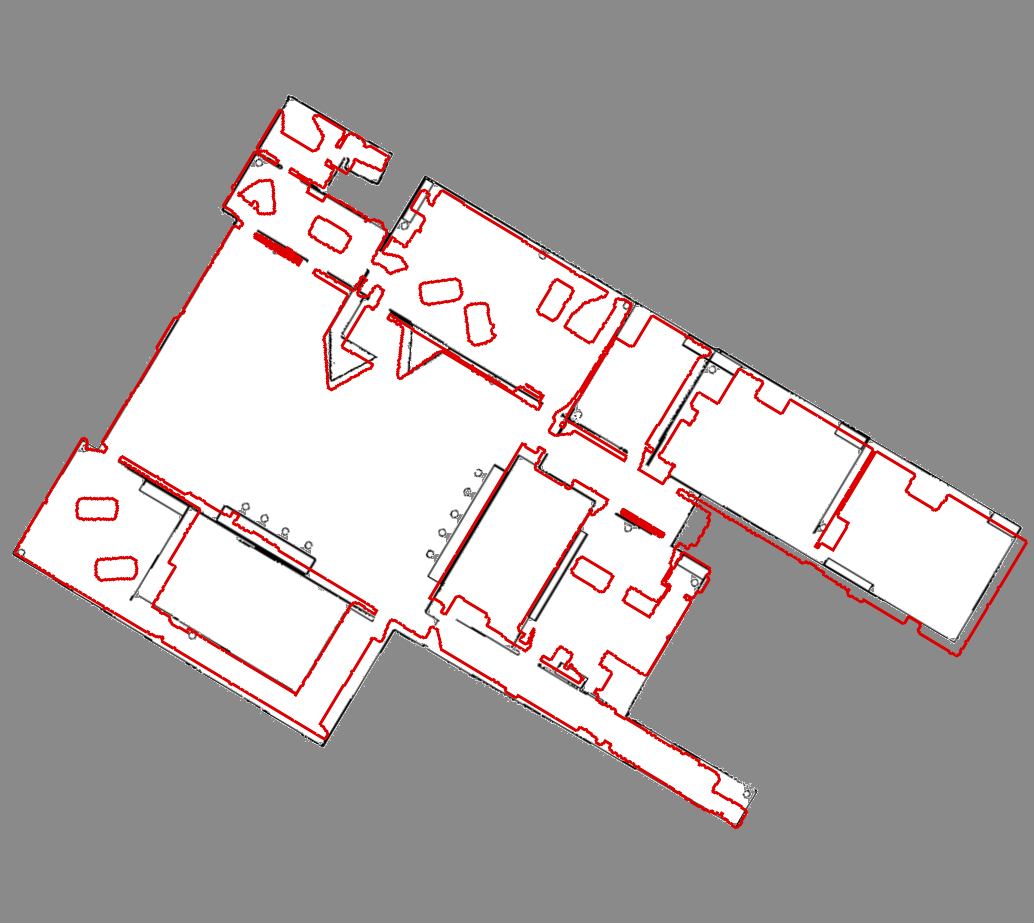}} \\
	\vspace{0.1cm}

	\rotatebox{90}{\hspace{1cm}lab\_f} 
	\subfigure[source image]{\includegraphics[width=0.15\linewidth]{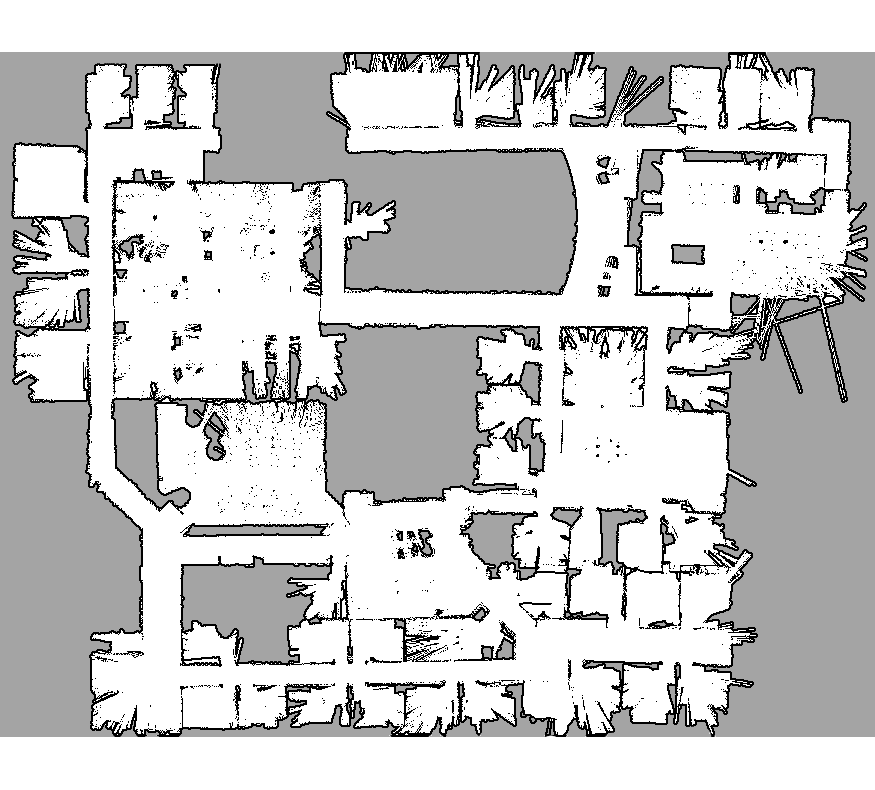}} \hspace{0.3cm}
	\subfigure[destination image]{\includegraphics[width=0.15\linewidth]{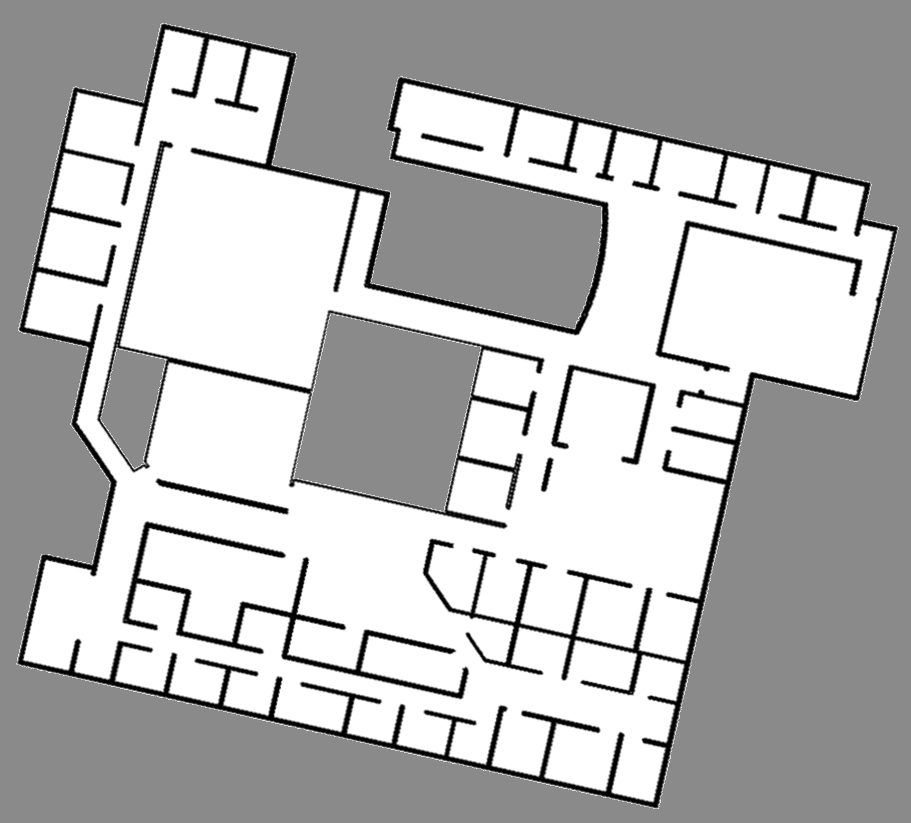}} \hspace{0.3cm}
	\subfigure[Shahbandi et al. \cite{shahbandi20192d}]{\includegraphics[width=0.16\linewidth]{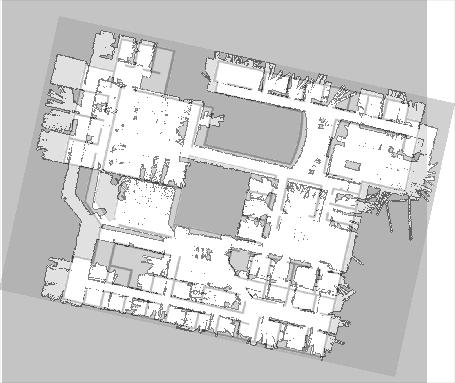}} \hspace{0.3cm}
	\subfigure[Ours]{\includegraphics[width=0.16\linewidth]{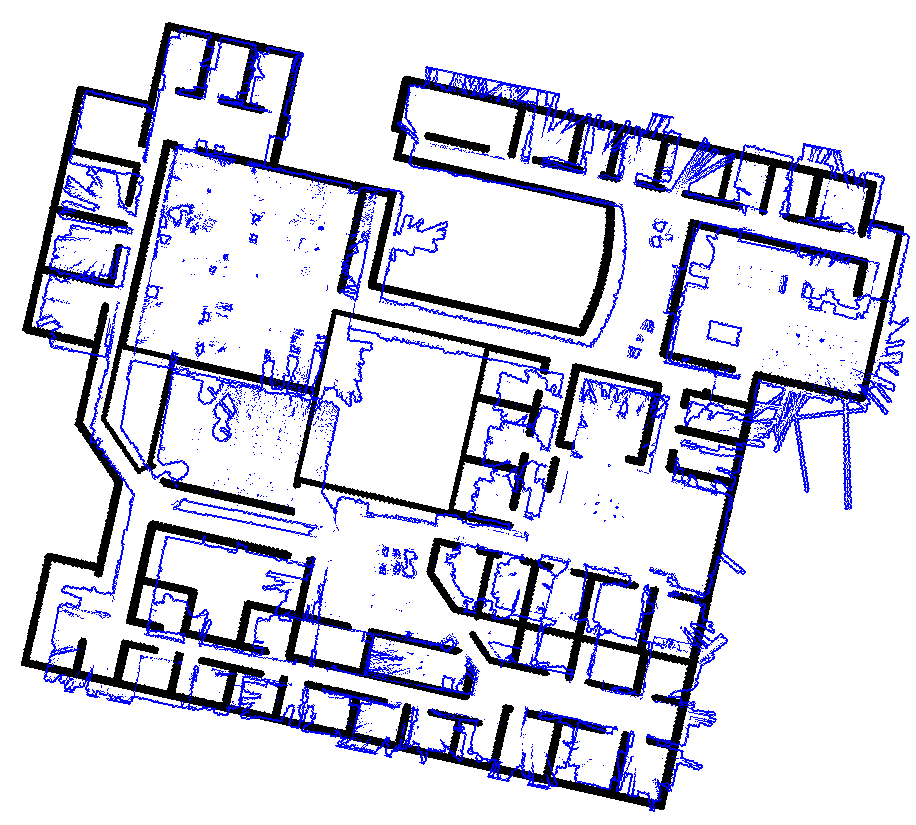}} \hspace{0.3cm}
	\subfigure[Ground truth]{\includegraphics[width=0.15\linewidth]{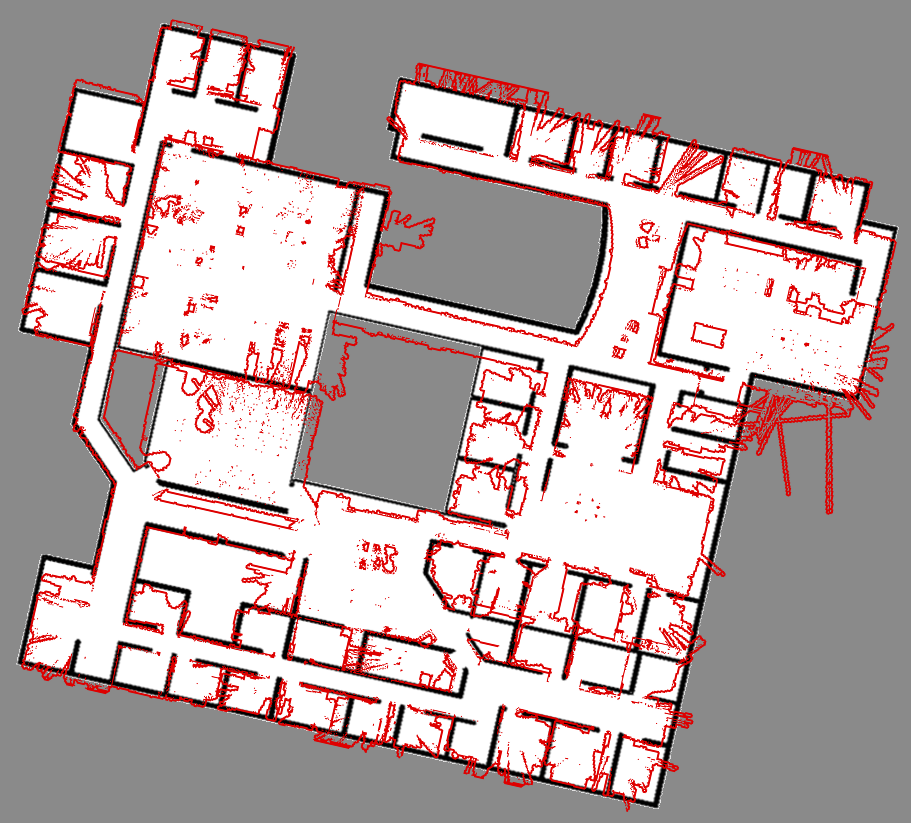}}
	
	\caption{Experiment result to show the performance of our algorithm. The source maps are occupancy grid maps which were generated by 2D SLAM algorithms. The destination map is a handmade layout image. And the ground truth is generated by manual alignment. The result shows that our algorithm has a better performance compared to the state-of-the-art method. \label{fig:results}}
\end{figure*}

In this experiment, we compared the methods on Bormann's dataset \cite{bormann2016room} with layout or artificial maps and robot maps, where one of the map in a pair is transformed so that we have rotation to estimate. The matching result of the two methods are compared with the ground truth are shown in Fig. \ref{fig:results}. Besides, we also compare the computation time of the two methods and  present the results in Table~\ref{tab:time}. All the experiments are finished on a PC with a Intel Core i7-6700 CPU (3.40GHz $ \times $ 8) and 16 GiB memory. The single threaded C++ code is running on the Ubuntu 16.04 operating system. The computation speed experiments provide separate analysis for the segmentation/ Area Graph generation time, matching time and total run time in the dataset. 
Using the Area Graph generation to segment the maps, the parameter width $w (m)$ to segment the different maps was set to $\{ 1.8, ~2.5, ~1.5, ~2.3, ~2.1, ~1.8 \}$. 
Though our method can deal with the maps with different scales by inputting resolutions, the other method is not good at this. We scale the input to the same scaling.

Since our matching is effected by the initial sample of the clustering algorithm, for each pair of input maps, we run our matching algorithm twenty times and recorded the correct match rate, as shown in Table \ref{tab:time}. 
One can observe that our method achieves a higher correctness. 
It can be seen that the other method has serious rotation errors: 180$\degree$ in the first map, 90$\degree$ in the second and third map. 
Also, our method shows a faster computation speed for segmentation and matching for all maps. 
Due to the reason that Shahbandi et al. don't use any geometric feature of the regions to match corresponding regions between maps, but estimate alignment transformation with the ''Least-squares estimation'' method for each pairwise regions' Oriented Minimum Bounding Boxes, about 90\% of the hypotheses are rejected. This process leads to 
long computation times for large maps. Therefore, our method performs better on large or complex maps.

\section{Conclusions}
\label{sec:conclusion}
In this paper, we proposed a fast 2D map matching method based on area feature matching, where we utilized an important characteristic of indoor architecture, that the location of doors and intersections is always fixed, as a novel area matching feature: passage distance. Other features are the area size and the convex hull longest length. We use the weighted sum of those features to find the best matches between areas. Using the overlap of good matched area candidates we finally extract the correctly matched areas and their transformation. The experiment compares our method with another area matching based. The experiments show that our method performs better on maps with a large quantity of rooms, which are normal in real buildings. Our algorithm also has a faster computation time. 

As future work we will apply our algorithm to map evaluation.

\renewcommand\arraystretch{1.5}
\begin{table*}[t]
	\centering
	\caption{The comparative analyses of computation times between two methods.}
	\begin{adjustbox}{width=1\textwidth}
		\begin{tabular}{@{}c|cc|cc|cc|cc|cc|cc|cc@{}}
			\toprule \toprule
			& \multicolumn{2}{c|}{freiburg 101} & \multicolumn{2}{c|}{intel} & \multicolumn{2}{c|}{lab\_a} & \multicolumn{2}{c|}{lab\_c} & \multicolumn{2}{c|}{lab\_d} & \multicolumn{2}{c|}{lab\_f} & \multicolumn{2}{c}{Average} \\
			& Shahbandi et al. & Ours & Shahbandi et al. & Ours & Shahbandi et al. & Ours & Shahbandi et al. & Ours & Shahbandi et al. & Ours & Shahbandi et al. & Ours & Shahbandi et al. & Ours \\ \midrule
			Segmentation/Arrangement Time (s) & 4.77 & \textbf{1.37} & 9.15 & \textbf{5.91} & 8.75 & \textbf{8.21} & 5.36 & \textbf{2.77} & 15.23 & \textbf{7.97} & 25.20 & \textbf{12.65} & 11.41 & \textbf{6.48} \\ \midrule
			Matching Time (s) & 1.41 & \textbf{0.004} & 3.85 & \textbf{0.085} & 4.56 & \textbf{1.715} & 3.05 & \textbf{0.069} & 18.91 & \textbf{0.049} & 200.96 & \textbf{0.192} & 38.79 & \textbf{0.352} \\ \midrule
			Total Times (s) & 9.23 & \textbf{1.39} & 15.93 & \textbf{6.00} & 16.43 & \textbf{9.93} & 11.10 & \textbf{2.84} & 39.74 & \textbf{8.02} & 231.79 & \textbf{12.85} & 54.04 & \textbf{15.32} \\ \midrule
			Correctness (\%) & 0 & \textbf{100} & 0 & \textbf{90} & 0 & \textbf{100} &  \textbf{100} & \textbf{100} &  \textbf{100} & \textbf{100} &  \textbf{100} & 95 & 50 & \textbf{97.5} \\	 \bottomrule \bottomrule\hline
		\end{tabular}
	\end{adjustbox}
	\label{tab:time}
\end{table*}

\IEEEtriggeratref{4}

\bibliographystyle{IEEEtran}
\bibliography{bibliography.bib}

\end{document}